\documentclass[twocolumn]{svjour3}           
\smartqed  

\input{header}

\journalname{The Visual Computer}

\begin{document}

\title{An effective and friendly tool for seed image analysis}


\author{Loddo A. \and
	Di Ruberto C. \and
	Vale A.M.P.G. \and
	Ucchesu M. \and
	Soares J.M. \and
	Bacchetta G.
}


\institute{Loddo A., Di Ruberto C.\at
	Department of Mathematics and Computer Science, University of Cagliari, \\
	via Ospedale 72, 09124 Cagliari, Italy. \\
	\email{andrea.loddo@unica.it}
	\and
	Vale A.M.P.G.\at
	Escola Agrícola de Jundiaí (EAJ), Universidade Federal do Rio Grande do Norte (UFRN), \\
	Rodovia RN 160, Km 03, CEP 59280-000 Macaíba (RN), Brasil.
	\and
	Ucchesu M., Bacchetta G. \at
	Centro Conservazione Biodiversità (CCB), Dipartimento di Scienze della Vita e dell’Ambiente (DiSVA), Università degli Studi di Cagliari, \\
	Viale S. Ignazio da Laconi 13, 09123 Cagliari, Italy.
	\and
	Soares J.M. \at
	Universidade Federal do Rio Grande do Norte (UFRN), \\
	Rodovia RN 160, Km 03, CEP 59280-000 Macaíba (RN), Brasil.
	\and
	Bacchetta G. \at
	Hortus Botanicus Karalitanus (HBK), Università degli Studi di Cagliari, \\
	Viale S. Ignazio da Laconi 11, 09123 Cagliari, Italy.	
}

\date{Received: date / Accepted: date}

\maketitle

\begin{abstract}
	Image analysis is an essential field for several topics in the life sciences, such as biology or botany. In particular, the analysis of seeds (e.g. fossil research) can provide significant information on their evolution, the history of agriculture, plant domestication and knowledge of diets in ancient times. 
	
	This work aims to present software that performs image analysis for feature extraction and classification from images containing seeds through a novel and unique framework. In detail, we propose two plugins \emph{ImageJ}, one able to extract morphological, textual and colour features from seed images, and another to classify seeds into categories using the extracted features. The experimental results demonstrated the correctness and validity of both the extracted features and the classification predictions. The proposed tool is easily extendable to other fields of image analysis.
\end{abstract}


\section{Introduction}
\label{sec:intro}
Due to its wide range of applications, image analysis plays an important role in the field of life sciences. Image analysis and image processing methods have become essential for understanding various medical features or performing meaningful quantitative measurements on image objects. Haematology (\cite{Dirub2020}, \cite{Remeseiro2021}), biology (\cite{Campanile19}, \cite{Ahmad2021}), and botany (\cite{bohl2015modeling}, \cite{LoBianco2017a}, \cite{hu2017data}) are a few examples of application fields in this context. 
In particular, image analysis techniques have become more reliable with the development of fluorescence and high-resolution microscopes, gaining the interest of biologists. The ability to study in-depth the structural details of biological elements, such as organisms and their parts, can profoundly impact biological research. In fact, thanks to these advances, several works have been carried out in the field of plants, related to the characterisation of germplasm (\cite{Bacchetta2011a}, \cite{Frigau2019}, \cite{LoBianco2015}, \cite{LoBianco2017a}, \cite{LoBianco2017b}), the identification of plant seed materials unknown in archaeobotany (\cite{Bouby2013}, \cite{Orru2013}, \cite{Terral2010}, \cite{Sabato2015}, \cite{Ucc2015}, \cite{Ucc2016}, \cite{Ucc2017}), or in agronomy to distinguish and group cultivars (\cite{Piras2016}, \cite{Sarigu2017}, \cite{Sau2018}, \cite{Sau2019}). According to~\cite{Kamilaris}, image analysis is an important area of research in agriculture for image classification, anomaly or disease detection, and several related tasks.
Plant science is the main subject of this work; more specifically, it concerns carpology, which is the discipline that studies the seeds and fruits of spermatophytes from both a morphological and structural point of view. This is of fundamental importance, for example, for palaeobotany, palaeoenvironmental studies and ecology applied to the remains of the past (palaeocarpology). The operators acquire the images through a digital camera or a flatbed scanner. Compared to the digital camera, the flatbed scanner offers the advantage of consistent illumination and a known image scale, commonly expressed in dots per inch (DPI), providing quality and speed of execution of the workflow (\cite{Lind2012}). This work presents new software for extracting features from biological organisms, particularly their images and, specifically, seeds.
Nowadays, one of the most used tools by biologists is certainly~\cite{ImageJ} (\cite{Landini}, \cite{Lind2012}). It is defined as one of the standard image analysis software, as it is freely available, platform-independent and applicable by biological researchers to quantify laboratory tests.
Compared to manual analysis, the use of seed image analysis techniques brings several advantages to the process:
\begin{enumerate*}
    \item it speeds up the analysis process;
    \item it minimises distortions created by natural light and microscopes;
    \item it automatically identifies specific features;
    \item it automatically classifies families or genera.
\end{enumerate*}

To achieve these benefits, a classical image analysis process employs a four-step workflow: preprocessing, segmentation, feature extraction and classification (\cite{Gonz2018}), although, since the explosion of the AlexNet convolutional neural network (CNN) (\cite{Krizhevsky2012}), a new approach has emerged. CNN-based workflows, and deep learning workflows in general, do not use the traditional image analysis workflow because CNNs can extract features independently without explicitly using feature descriptors or dedicated feature extraction strategies. Regarding the traditional approach, image preprocessing techniques prepare the image before analysing it to eliminate possible distortions or unnecessary data or highlight and enhance some important features for further processing. Segmentation divides significant regions into sets of pixels with common characteristics such as colour, intensity or texture. The purpose of segmentation is to simplify and change the image representation into something more meaningful and easier to analyse. Extracting features from the regions of interest identified by segmentation is the next step. Features can be based on shape, structure or colour (\cite{Dirub2015}, \cite{Dirub_2009}). The final step is classification, i.e. the association of a label with the object under investigation using supervised or unsupervised machine learning methods. 

In this paper, we focus on feature extraction and classification steps from both traditional and CNN-based perspectives.
In detail, our contribution is fivefold: 
\begin{enumerate*}
    \item we propose a new open-source tool for extracting features from seed images;
    \item we propose a new open-source tool for feature classification based on four different models;
    \item we compared the classification accuracy of four different models and different features combination;
    \item we compared traditional classification models with some classical CNNs;
    \item we propose a unique framework to analyse seed images turned to be effective, fast and mostly easy to use for non-expert operators.
\end{enumerate*}

Our work aims to classify single seeds belonging to the same family, where differences in colour, shape and texture may be much more imperceptible. We also want to highlight how traditional classification methods trained with handcrafted (HC) features can bring greater speed and accuracy close to CNNs in this task.

The rest of the paper is organised as follows. Section~\ref{sec:related} presents state of the art on plant science works, with a particular focus on seeds image analysis. Section~\ref{sec:materials_methods} presents the dataset employed and introduces the ImageJ environment used to implement our proposed plugins. They are described in~\ref{sec:tool}. The experimental evaluation and the used dataset are discussed in Section~\ref{sec:results}, and, finally, in Section~\ref{sec:conclusions} we give the conclusions of the work.


\section{Related works}
\label{sec:related}
The analysis of seeds or leaves for different purposes is one of the possible processes of biological image analysis in the field of plant science. Both traditional (\cite{Campanile19}, \cite{DiRub2014}, \cite{Put16}) and deep learning (\cite{Zhang2019}, \cite{Loddo2021}) approaches have been widely exploited in this context. For example, the analysis of seed fossils can provide important information about their evolution, the origin of agriculture, the process of domestication and knowledge about diets in ancient times. Usually, such fossils are preserved through a carbonisation process to avoid a microbial attack.
Regarding traditional approaches, in \cite{Ucc2016} Ucchesu et al. carried out several charring experiments to reproduce the same burning conditions as in archaeological contexts in order to compare some seeds present today in Sardinia, Italy, with presumed archaeological fossils and classify them as such.
In \cite{Ucc2015} the authors performed a morphological comparison between archaeological seeds and recent wild seeds. This work showed that archaeological seeds have significantly similar characteristics to modern seeds. Sabato et al. in \cite{Sabato2015} analysed the genetic, morphological and colourimetric differences of 124 seed types of \emph{Cucumismelo}, from 48 different countries. The morpho-colourimetric analysis revealed two subspecies of melons and also identified six different varieties. The work of Orr\'u et al. (\cite{Orru2012}) is the first attempt to validate a morpho-colourimetric method by direct comparison with molecular data from germ plasma, showing that the 113 proposed characteristics are adequate to discriminate similar groups. Lo Bianco et al. (\cite{LoBianco2015}) identified 67 different Italian bean types (\emph{Phaseolusvulgaris L}) using morphological traits. A total of 138 descriptors, including shape and texture, were extracted from each seed using image analysis techniques. Through linear discriminant analysis (LDA), the authors performed a comparative analysis to test the possibility of distinguishing between seeds from the same soil but grown with different agricultural practices. Initially, it was possible to discriminate three main seed categories with an accuracy of 99.1\%. An automatic feature extraction scheme for plant species was proposed by \cite{Campanile19}, while \cite{DiRub2014} and \cite{Put16} implemented a leaf recognition system and a leaf detection system, respectively. The former is implemented with a Support Vector Machine (SVM) for plant classification based on leaf images' shape, colour, and texture characteristics. The latter can automatically detect the leaf of interest through a segmentation phase based on the extraction of saliency maps as a starting point for a region growing strategy. 

Regarding deep learning strategies, \cite{Hall2015} proposed a CNN to address leaf classification of different plant species, using a dataset of over 1,900 images divided into 32 species. Similarly, in~\cite{Slado2016}, CaffeNet is used to identify leaf diseases in a database containing about 4,500 images, while Zhu et al.~\cite{Zhu2021} aimed to recognise the appearance quality of carrots by training a Support Vector Machine classifier with deep features, with excellent results from the ResNet101 network.

Deep learning is also used for identification and classification; in particular, AlexNet and GoogleNet are used to identify 14 crop species and 26 diseases in~\cite{Moha2016}, and LeNet is used for diseased banana leaves recognition in ~\cite{Amara2017}. \cite{Junos2021} proposed a system for detecting loose palm fruits from images acquired under various natural conditions. It was based on an improved version of YOLOv3. They also made a new dataset of oil palm loose fruits acquired by an unmanned aerial vehicle (UAV) and a mobile camera. \cite{Gajjar2021} realised a comprehensive framework for real-time identification of diseases in a crop. They proposed a novel CNN architecture to identify and classify 20 different healthy and diseased leaves of 4 different plants. It achieved an accuracy of 96.88\%, higher than the accuracy achieved by existing architectures.
In~\cite{Kussul2017}, the research focused on the classification of wheat, maise, soybean, sunflower and sugar beet crops using CNNs produced by the authors. In~\cite{Mort2016}, the authors used a modified version of VGG16 to identify oilseed crops, radishes, barley, grass and weeds. Rebetez in~\cite{Rebe2016} classified various crop styles from drone images using CNN and HistNN (an RGB histogram). 
\cite{Gulzar2020} dealt with seed classification using CNN and transfer learning, although the authors addressed the problem of seeds belonging to different phyla or classes starting from a set of seeds. Przybylo et al.~\cite{Przybylo19} focused on acorn classification based on the colour and image intensity of seed sections as a feature. The authors obtained an accuracy of 85\%, which is comparable to a manual assessment of oak seed viability using a CNN strategy. Furthermore, they studied the impact of various image representations (colour, entropy, edges) and the architecture of the network and its parameters on the classification results. Finally, \cite{Loddo2021} realised and proposed a brand new CNN architecture for seed image classification and retrieval tasks, called SeedNet. It achieved 97.47\% accuracy in the same dataset used in this study. However, as it can happen when using deep strategies, the running time was very high, especially if compared to traditional machine learning techniques. 

This work's contributions are an in-depth investigation of how HC features can improve classification results if several features, even heterogeneous ones, are fused together. For the purpose of this study, no feature selection schemes were investigated to preserve the original nature of the combined features. Feature fusion is a strategy deeply studied in machine learning related works. For example, \cite{jing2014saliency} proposed a saliency detection model using the integrated features obtained from a feature fusion method that combined local and global saliency measures during the saliency calculation process. Their experiments showed that this model performs better than the ones trained with non-combined features. \cite{peng2015linear} realised a method for feature fusion of multimodal finger biometrics, in which linear discriminant multi-set canonical correlation analysis (LDMCCA) was adopted to combine multiple feature sets in a meaningful way. LDMCCA can preserve the intrinsic relationship within-class and between-class in unimodal features and distinctly express the discriminant correlation between multimodal features. In addition, it can further perform feature reduction alone on fused features. The LDMCCA approach significantly improves existing fusion approaches. \cite{gad2019iot} proposed a connection between client and server based on an iris recognition system. The system obtained an accuracy of 99.86\% with a high recognition speed, using a fusion of textural and statistical iris features.


\section{Material and Methods}
\label{sec:materials_methods}
In this work, we exploited a local dataset~\ref{sec:dataset}. It contains heterogeneous seed images, both in number and characteristics and is publicly available on request.
The dataset has been preprocessed as depicted in Sec.~\ref{sec:seeds_analyser} and used for the classification of seeds families task.

\subsection{Dataset description}
\label{sec:dataset}
\begin{figure}[tbp]
	\centerline{\includegraphics[scale=0.25]{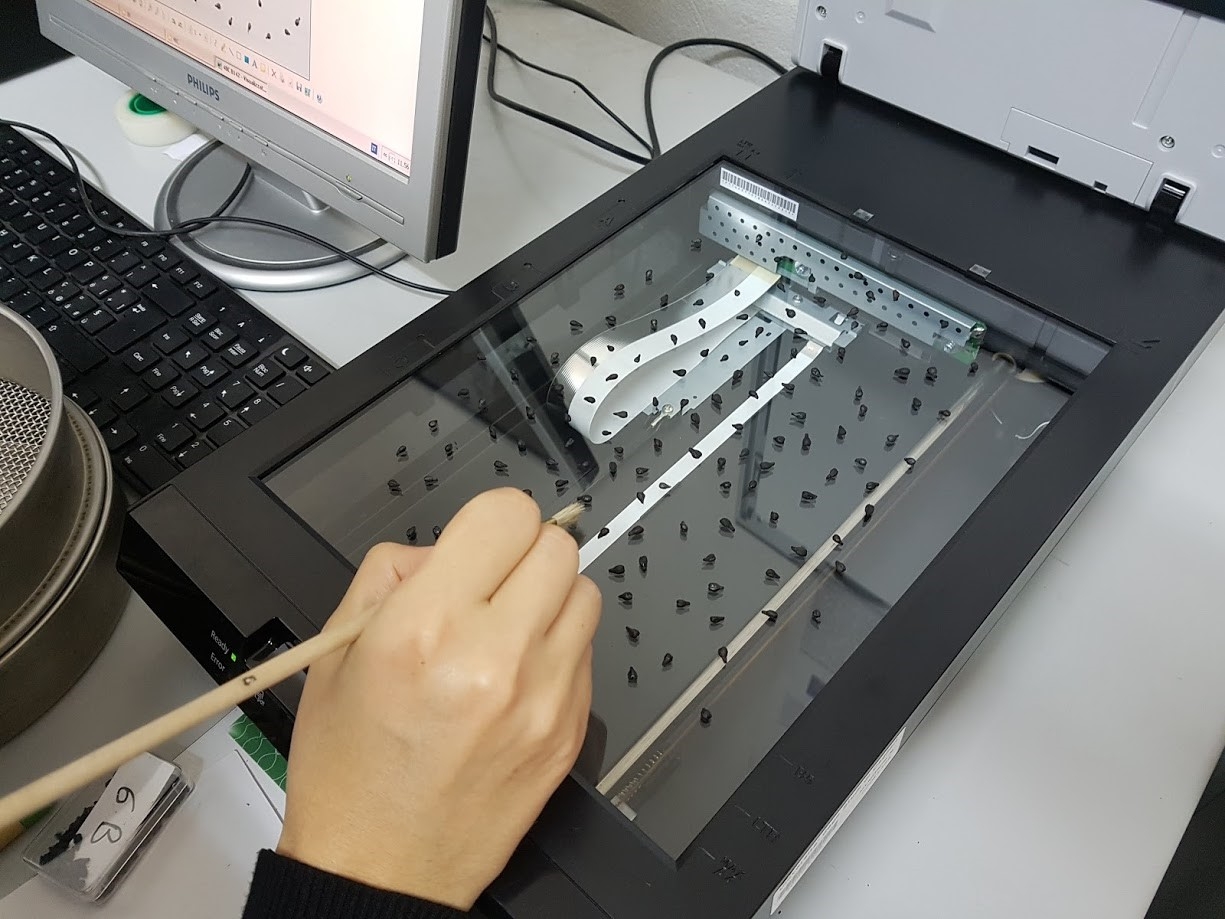}}
	\caption{Seeds acquisition on the flatbed scanner for the \emph{Fabaceae} dataset.}
	\label{Acquisition}
\end{figure}

For this study, we used an image database containing 3,386 samples of 120 plant species belonging to the \emph{Fabaceae} family.
We chose the \emph{Fabaceae} family because it is one of the most influential families and shows significant variability in the size and colour of their seeds. All samples come from the basic collection of the Banca del Germoplasma della Sardegna (BG-SAR), University of Cagliari, Italy.
During the acquisition, the operators arranged the seeds on the flat scanner, separating them to avoid overlapping. Then, the area occupied by the seeds was covered with a tray coated with a blue background for the digital image, as shown in Fig. \ref{Acquisition}. The acquisition process used a minimum resolution of 400 DPI, and the resulting image was saved in the Joint Photographic Experts Group (JPEG) format with a resolution of $2125\times2834$ (\cite{Loddo20}).

\begin{figure}[tbp]
	\centerline{\includegraphics[scale=0.5]{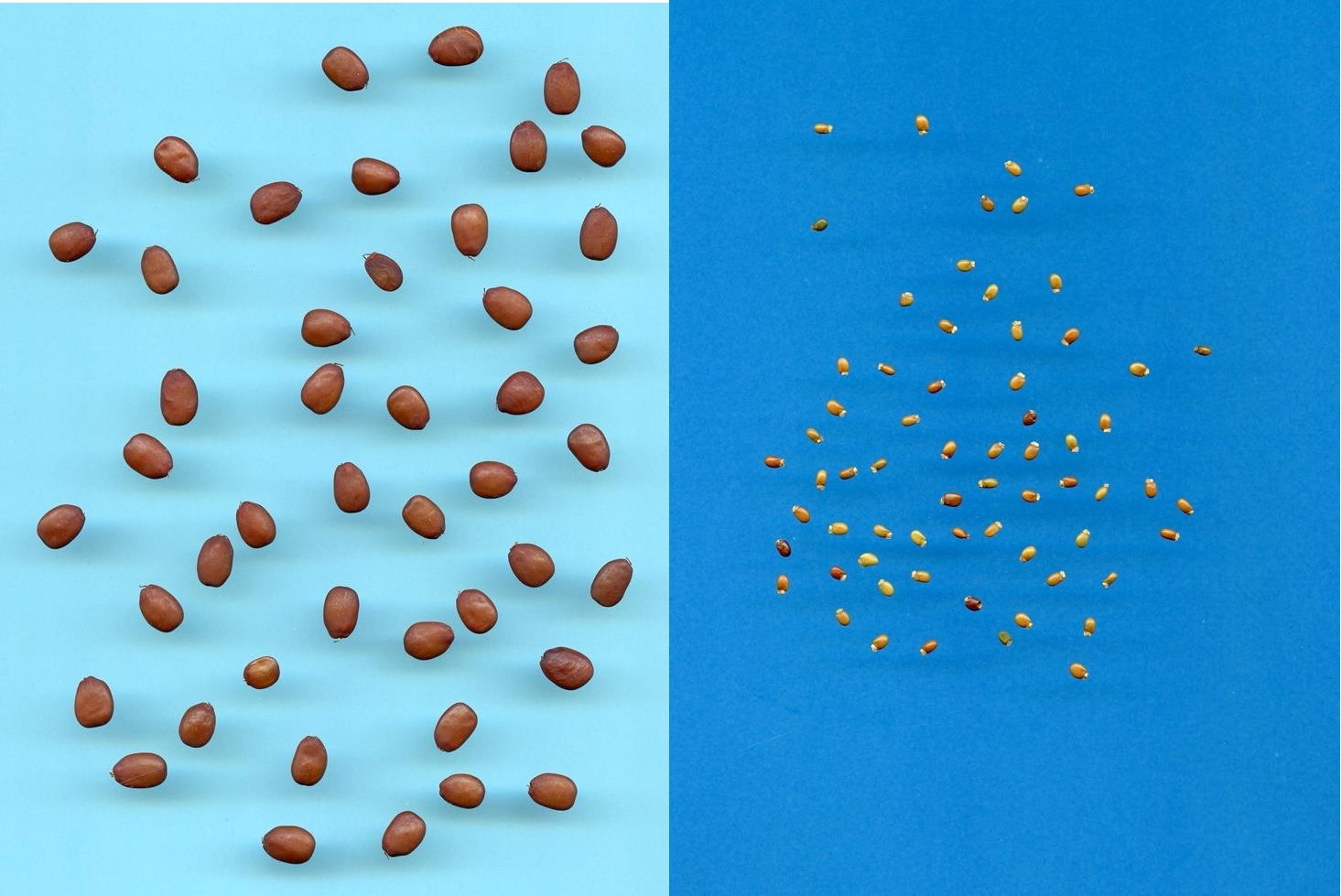}}
	\caption{Examples of seeds images present in the \emph{Fabaceae} dataset.}
	\label{Sardinia}
\end{figure}

For the classification task, we selected the more numerous species.
Next, we applied a preprocessing procedure. 
In particular, the collectors acquired these seeds' images on different backgrounds of various shades of blue. Consequently, it helps us to find the best images to extract crops of single seeds to classify. Fig. \ref{Sardinia} shows two sample images from the \emph{Fabaceae} database.

\subsection{ImageJ platform}
\label{sec:imagej}
ImageJ (\cite{ImageJ}) is open-source software for digital image processing, designed initially by Wayne Rasband from the National Institute of Health of the United States. It is written in Java language, and it runs as an online applet or offline with an installed Java Virtual Machine (JVM). The source code is publicly available and open to new contributions for the ImageJ community. ImageJ allows viewing, analysing, modifying, processing, saving, printing grayscale images (8-bit, 16-bit, and 32-bit) and colour images (8-bit and 24-bit); the supported formats are TIFF, JPEG, GIF, BMP, DICOM, FITS, and RAW. It was designed with an open architecture extendable with two kinds of extension, \emph{Java plugins}, and recordable \emph{macros}. Most of the existing plugins already permit to face of some image processing and analysis issues.
ImageJ allows the extraction of Region of Interest (ROI) pixel-wise or object-segmented statistics. It is possible to measure distances and angles, generate and visualise intensity histograms and draw profile lines (between defined points). ImageJ supports many standard image processing transformations, such as logical and arithmetic operations between images, brightness and contrast adjustment, convolution, Fourier analysis, smoothing, contour detection, median filtering, and mathematical morphology \cite{Gonz2018}. It is also possible to perform geometric transformations such as scaling, rotation, and reflection.

\begin{figure}[!t]
	\centerline{\includegraphics[scale=0.37]{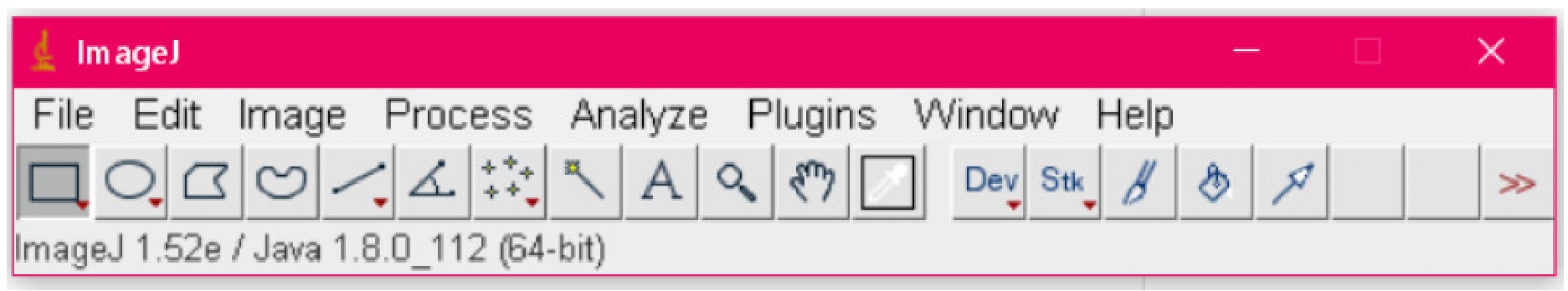}}
	\caption{ImageJ main interface.}
	\label{ImageJ_Interf}
\end{figure}

The interface of ImageJ is straightforward and intuitive, even for people without advanced computer and image analysis skills. 
The software is condensed in its menu bar that contains all the options, as shown in Fig. \ref{ImageJ_Interf}.

Basic operations, like file opening or simple editing options (File and Edit options) and more complex ones, such as segmentation operations, image enhancement, noise reduction, object counting, filtering, and other options ("Image," "Process," "Analyse" and "Plugins"), are accessible from the main menu.
Moreover, the toolbar contains several options to select regions or shapes in the image. 
The status bar shows the current pixel's coordinates and values when the cursor is over the image. Simultaneously, whether a filter operation is performed on the image, the status bar displays the execution time and the processing speed in $pixels / second$.
The progress bar, shown on the right-hand side of the status bar, shows the processing progress.

ImageJ allows an extension of its functionalities by additional components, like plugins and macros. The plugins require Java language, while macros require Java-like language. Plugins are generally faster and more flexible; on the other hand, macros are simpler to write and debug, but they are heavier than plugins computationally. It is the main reason that led plugins to play a fundamental role in functionality extensions and ImageJ since it implements a large part of its functionality with internal plugins.

\subsubsection{ImageJ's logical structure}
\label{sec:image_logical}
ImageJ presents a very extensive and complex class diagram.
\emph{IJ} is the main class of the framework: any processing starts from it, and consequently, all the other classes extend it or are part of \emph{ij}.* package. It takes an image in input and returns an object of the \emph{ImagePlus} type, ready for its analysis.
The created \emph{ImagePlus} object contains an object of the abstract \emph{ImageProcessor} class, which stores image data in 2D and provides methods for processing.
The \emph{GenericDialog} and \emph{ResultsTable} classes respectively manage input and output data. The first one allows the user to specify preferences and select options via several checkboxes, text boxes, and lists, while the second one shows the output results in tabular form. Moreover, the \emph{ROI} class permits processing image's objects.
This class includes several parameters and methods and is often associated with an object of type \emph{ImageStatistics} that consists of a series of measurements calculated on the ROI. The \emph{ROI} class uses the classes\emph{Polygon Java } and its subclasses to determine the points which constitute an area of interest. The \emph{Analyser} class returns an object based on particular options that analyse the image. An object of type \emph{ParticleAnalyzer} works in the same manner even though it analyses all the regions in an image one-by-one rather than the whole image. 
The class \emph{Histogram} provides an image histogram on a ROI. 
\emph{ChannelSplitter} returns a vector containing three \emph{ImagePlus}, each one corresponding to a single RGB (Red, Green, Blue) channels.
In the next section, we detail our proposed plugin structure, based on the \emph{ParticleAnalyzer}'s logical structure.


\section{A tool for seed image analysis}
\label{sec:tool}
ImageJ allows the realisation of new plugins, e.g., dealing with specific problems in the image analysis field or computing new features in different contexts.
Among the state-of-the-art, different plugins already exist and can extract features from seeds images, even though they are general-purpose. One of them is \emph{Particles8} \cite{Landini}, created by Gabriel Landini in 2005, with the last update in 2010. It provides the extraction of morphological features from binary images. Consequently, it does not provide the extraction of textural and colour features.
To simplify the botanists' seed analysis procedures, we followed their indications, and we realised a brand new plugin exclusively with \emph{ImageJ} classes without using any external plugins and increasing its extensibility. The proposed tool allows both the extraction and classification of features and can be used in many other application domains.

\subsection{A plugin for feature extraction from seeds image}
\label{sec:seeds_analyser}
\begin{figure}[htbp]
	\centering
	\includegraphics[scale=0.6]{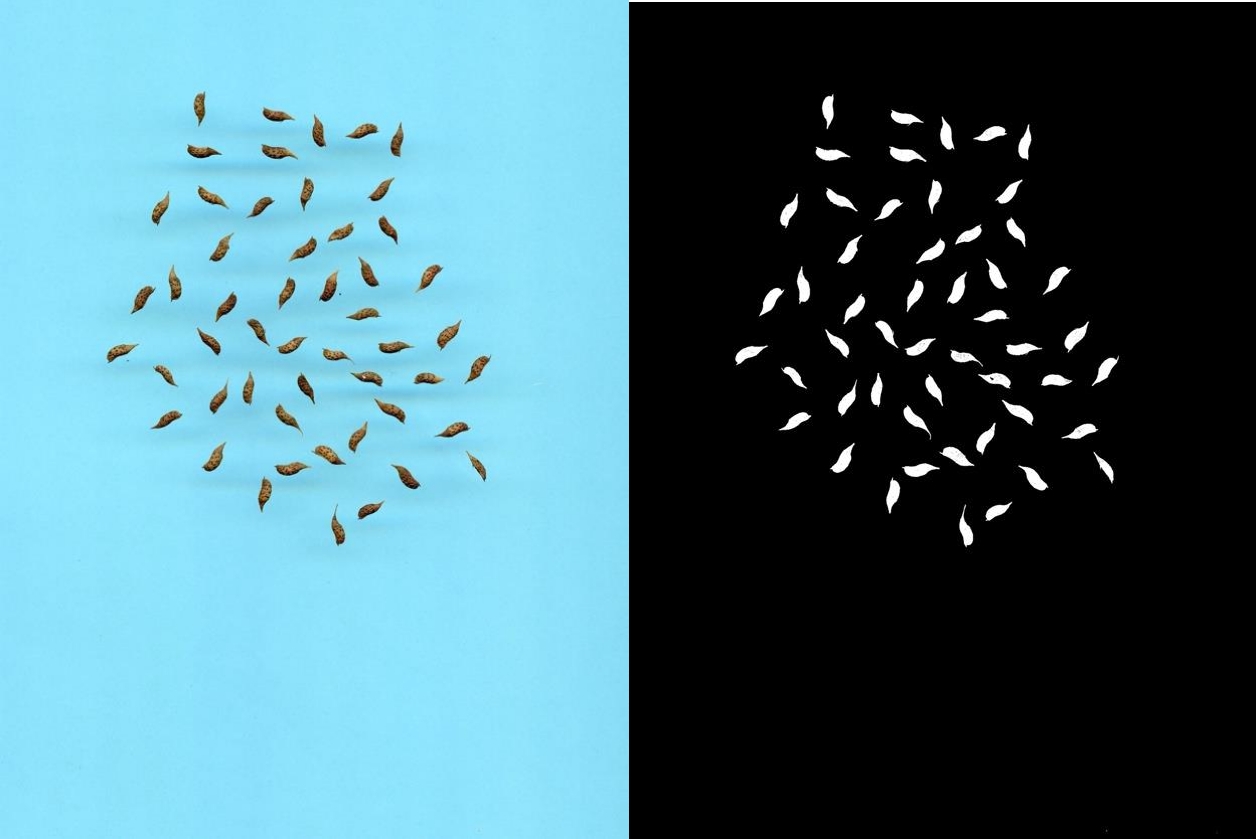}
	\caption{Example of an image from the \emph{Fabaceae} database and its derived binary mask.}
	\label{Sardinia_Masks}
\end{figure}

\begin{figure}[htbp]
	\centering
	\includegraphics[scale=0.4]{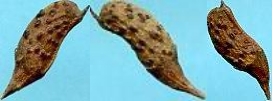}
	\caption{Some examples of seed images extracted from the image of Figure \ref{Sardinia_Masks} of the \emph{Fabaceae} database.}
	\label{Sardinia_Crops}
\end{figure}

\emph{SeedsAnalyser} is the plugin for feature extraction proposed in this work and needs a minimal number of user interactions. It aims to identify and analyse multiple seeds represented in a digital image. The input is a single image acquired with a quasi-uniform blue background.
The plugin is also able to analyse all the images in a specific folder at the same time. Each image can present a blue background in a wide range of tones, varying from a very light blue to a very dark navy blue. After the acquisition, the image is preprocessed to correctly separate the regions of interest, namely the single seeds. 
It is possible to isolate the Blue value of the RGB images to get pure masks as the dataset backgrounds are all in various shades of blue (see Fig.  \ref{Sardinia_Masks}), as indicated in our previous work \cite{Loddo20}. 
Creating a single seed dataset from the original database is possible thanks to different backgrounds of various blue shades. It allowed finding the best images to work on and making the binary masks to extract the single seeds by an automatic thresholding procedure.
During the acquisition, the seeds were well spaced from each other. Therefore the bounding box of each region allowed the creation of a single seed image for analysis quickly.
From the \emph{Fabaceae} dataset, we selected the images containing the most numerous samples per family for 23 different ones and nearly 2000 seeds.
We discarded some families due to the low sharpness and too small size of the seeds. Fig. \ref{Sardinia_Masks} and Fig. \ref{Sardinia_Crops} show an original sample image with its derived binary mask and some seed images extracted from it, respectively. 

\begin{figure*}[htbp]
	\centering
	\includegraphics[height=0.35\textheight]{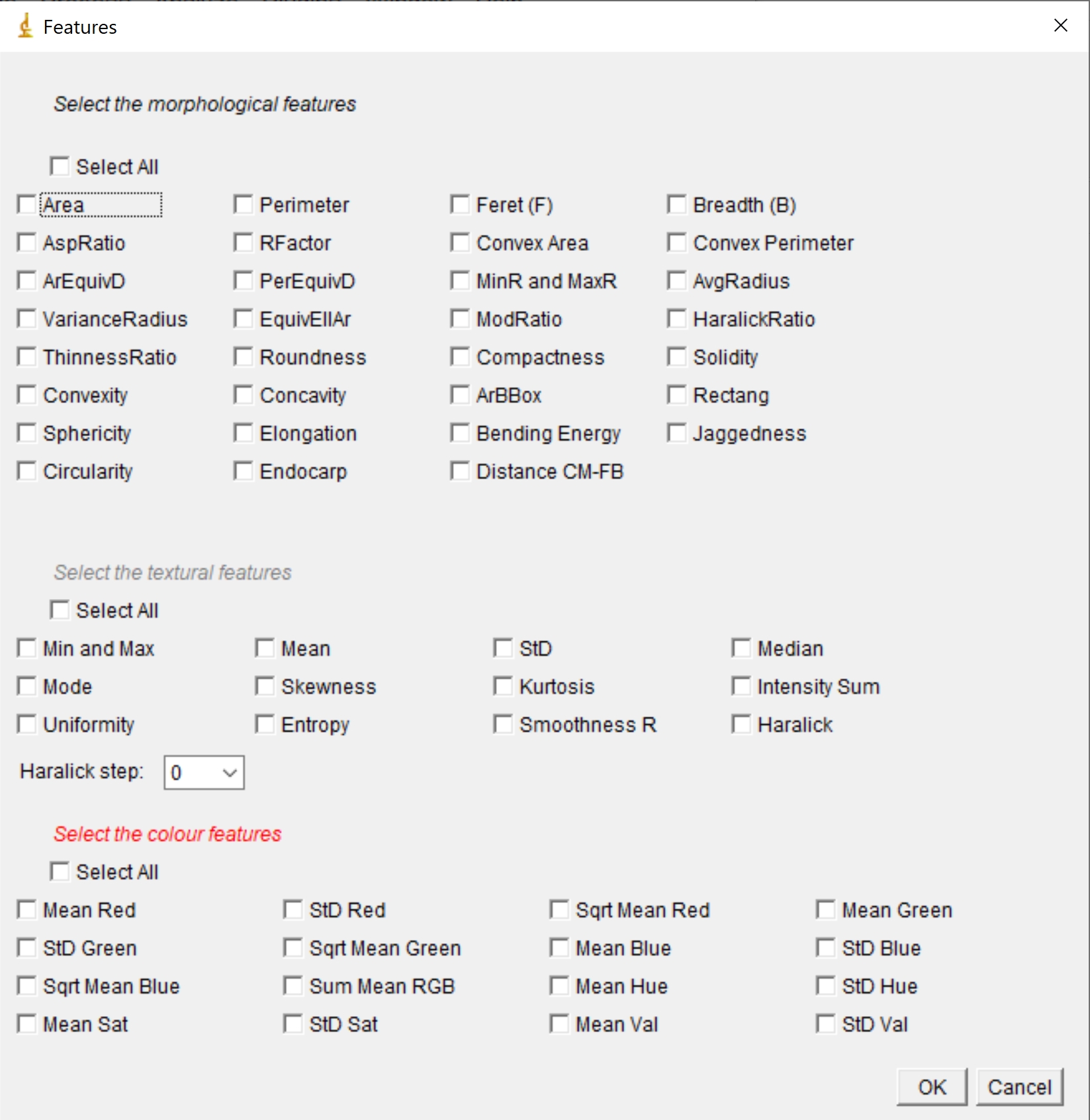}
	\caption{\label{fig:features_selection}Window for feature selection.}
\end{figure*}

Once the images have been preprocessed, i.e. segmented by automatic thresholding, and the unique image is ready to be analysed, the plugin requires the selection of some key parameters. They improve the research and detection of the regions of interest, specifically the minimum and maximum area size, measured in square pixels and, if wanted, a specific circularity of the objects, by default ranging from 0 to 1.
Finally, the users can choose the features of interest from the "Feature" window, as shown in Fig.  \ref{fig:features_selection}.

\emph{SeedsAnalyser} implements up to 64 features. In particular, 32 are morphological, 16 textural and 16 colour intensity values. It is crucial to notice that, among the texture features, Haralick's GLCM, which describes the pairwise arrangement of pixels with the same grey-level (\cite{Haralick73}), was used in this study to extract information of local similarities. They all permit their computation with the typical four different degrees: 0\degree, 45\degree, 90\degree, 135\degree. More precisely, we extracted the following second-order statistics from GLCM: energy, contrast, correlation and homogeneity.
ImageJ already contains a plugin that works in a way similar to \emph{Seeds Analyser}, even though it only offers 18 features, and it does not have a multi-image workflow.
To sum up, after the initial preprocessing step, our plugin can detect each single seed present in the original RGB image from which the user can select the morphological, textural and colour features to extract. Table \ref{table:morph} and Table \ref{table:gray} present the implemented morphological and textural features, respectively, and their relative descriptions, while Table \ref{table:col} describes the computed features of the RGB and HSV colour spaces.

\begin{table*}[htbp]
	\caption{Morphological features from binary image}
	\centering 
	\footnotesize
	\scalebox{0.90}{
		\begin{tabular}{ l l } 
			\hline
			\textbf{Feature} & \textbf{Description} \\
			\hline
			\emph{Area} & Seed area (in pixels) \\
			\emph{Perimeter} & Length of the seed contour \\
			\emph{Feret} & Longest traceable diameter with two points of the seed's outline as endpoints, called Lenght\\
			\emph{Breath} & Length of longest traceable axis perpendicular to the Feret, also called Width\\
			\textit{AspRatio} & \textit{Feret}/\textit{Breadth}, also called eccentricity or rectangularity ratio\\
			\emph{ConvexArea} & Area of the convex polygon drawn between the external points of the region\\
			\emph{ConvexPerimeter} & Perimeter of the convex polygon \\
			\textit{RFactor} & Shape factor, defined as $CovenxArea/(Feret\times pi)$\\
			\textit{ArEquivD} & Diameter of the circle with equivalent area of the region, defined as $\sqrt{4/\pi \times Area}$  \\
			\textit{PerEquivD} & Diameter of the circle having the same perimeter of the region, $Area/\pi$ \\
			\textit{MinR} and \textit{MaxR} & Radii of the inscribed and the enclosing circles centred at the center of mass\\
			\textit{AvgRadius} & Average length of the radii calculated starting from the center of mass \\
			\textit{VarianceRadius} & Variance of radii \\ 
			\textit{EquivEllAr} & Area of the ellipse having \emph{Feret} and \emph{Breadth} as axes \\
			\textit{Modification Ratio} & Shape measure, defined as $(2 \times MinR)/Feret$ \\
			\textit{Haralick Ratio} & Ratio between the average and the standard deviation of the radii \\
			\textit{ThinnessR} & Thinness Ratio, also called shape, given by $ Perimeter^2 / Area $ \\
			\textit{Roundness} & Measure of roundness, defined as $4 \times Area/(\pi \times Feret^2)$ \\
			\textit{Compactness} & Measure of compactness, expressed by $\sqrt{4/\pi \times Area/Feret}$  \\
			\textit{Solidity} & Measure of solidity, defined as $Area/ConvexArea$ \\
			\textit{Convexity} & Measure of convexity, also called roughness, defined as $ConvexPerimeter/Perimeter$ \\
			\textit{Concavity} & Measure of concavity, defined as $ConvexArea$ - $Area$\\
			\textit{ArBBox} & Area of the bounding box containing the region \\
			\textit{Rectangularity} & Also called extent, defined as $Area / ArBBox$ \\
			\textit{Sphericity} & Also called radius ratio, expressed by $MinR/MaxR$ \\
			\textit{Elongation} & Inverse of the circularity, defined as $Perim^2 / (4 \times \pi \times Area) $\\
			\textit{Bending Energy} & Defined as the sum of the squared curvature along the entire contour \\
			\textit{Jaggedness} & Measure representing if a seed is "serrated", defined as $(2\times \sqrt{\pi\times Area})/Perimeter$ \\
			\textit{Circularity} & Also called shape factor, obtained by $2\times \pi \times Area/Perimeter^2$ \\
			\textit{Endocarp} & Number of pixels forming the seed endocarp\\
			\textit{FBtoCM} & Distance between the intersection coordinates of seed length and width and center of mass \\
			\hline
	\end{tabular}}
	\label{table:morph}
\end{table*}

\begin{table*}[htbp]
	\caption{Texture features from grayscale image}
	\centering 
	\footnotesize
	\scalebox{0.95}{
		\begin{tabular}{ l l } 
			\hline
			\textbf{Feature} & \textbf{Description} \\
			\hline
			\textit{Min} and \textit{Max} & Minimum and maximum gray value in the region \\
			\textit{Mean} & Average gray value in the region\\
			\textit{StD} & Intensity standard deviation as contrast measure \\
			\textit{Median} & Median of the gray values \\
			\textit{Mode}  & Mode of the gray values \\
			\textit{Skewness} & Measure of the symmetry of the graylevel histogram around the average value \\
			\textit{Kurtosis} & Measure of the "tailedness" of the graylevel histogram \\
			\textit{Intensity Sum} & Sum of the gray values of the region \\
			\textit{Uniformity} & Maximum when all the gray levels in the histogram are equal \\
			\textit{Entropy} & Measure of variability of grey level distribution \\
			\textit{Smoothness R }& Measure of smoothness \\ 
			\textit{Haralick} & GLCM's computed second-order statistics (Energy, Contrast, Correlation, Homogeneity) \\
			\hline
	\end{tabular}}
	\label{table:gray}
\end{table*}

\begin{table*}[htbp]
	\caption{Color features from RGB and HSV color spaces}
	\centering 
	\footnotesize
	\begin{tabular}{ l l } 
		\hline
		\textbf{Feature} & \textbf{Description} \\
		\hline
		\textit{Mean Red (MR) }& Average of Red channel values \\
		\textit{StD Red} & Standard deviation of Red channel values \\
		\textit{SqrtMean Red} & Square root of the mean value for Red channel  \\
		\textit{Mean Green (MG)} & Average of Green channel values \\
		\textit{StD Green} & Standard deviation of Green channel values \\
		\textit{SqrtMean Green} & Square root of the mean value for Green channel \\
		\textit{Mean Blue (MB)} & Average of Blue channel values \\
		\textit{StD Blue} & Standard deviation of Blue channel values \\
		\textit{SqrtMean Blue} & Square root of the mean value for Blue channel \\
		\textit{Mean RGB} & \( \frac{MR + MG + MB}{3} \)\  \\ 
		\textit{Mean Hue} & Average tone of Hue channel \\
		\textit{StD Hue} & Standard deviation of Hue channel values \\			
		\textit{Mean Sat} & Average tone of Saturation channel \\
		\textit{StD Sat} & Standard deviation of Saturation channel values \\
		\textit{Mean Val} & Average tone of Value channel \\
		\textit{StD Val} & Standard deviation of Value channel values \\
		\hline
	\end{tabular}
	\label{table:col}
\end{table*}

\subsection{A plugin for feature classification}
\label{sec:feature_classifier}
Up to now, we have obtained the features for each seed from \emph{SeedsAnalyser} \ref{sec:seeds_analyser}, and they can now be fed to the classification plugin, called \emph{SeedsClassifier}.
It offers four different classifiers, namely kNN, Naive Bayes, Random Forest and SVM. Weka \cite{Weka} includes all of them; therefore, they can be imported individually from their respective packages. All classifiers belong to their java class, where they implement the Classifier interface responsible for defining and realising the classification procedure.
At the plugin's start, the user can choose whether to load an existing model or start a new training phase on new data. If the user chooses to proceed from scratch, i.e. also with the training phase, the user will be asked to enter the ARFF file's name with the training dataset. For practical reasons, this file must be located in the main folder of the framework. The predictions will be displayed in a window called Predictions.
As we mentioned earlier, \emph{SeedsClassifier} plugin allows for the use of four classification algorithms. We now briefly describe how they operate and differ from each other.
In general, Naive Bayes classifiers are probabilistic models that use Bayes' theorem with strict independence assumptions between the features. 
KNN uses the \textit{k} closest training samples in the dataset as input and then uses a neighbour voting strategy to rank and classify new objects. Generally speaking, the larger \emph{k} is, the more the noise associated with classification is reduced, but class recognition becomes more difficult. 
The Support Vector Machine (SVM) is a non-probabilistic binary linear classifier that categorises objects by mapping examples to points in space to maximise the width of the distance between categories.
Finally, Random Forest is made up of many individual decision trees that work together to form an ensemble. Each tree predicts a class, and the class with the most votes is the model prediction. However, there is a need for each tree not to correlate with the others. This would jeopardise the classifier's final decision. In this way, the trees protect each other from their errors.
Given our feature spaces' abundance and diversity, we choose these classifiers to ensure classification accuracy, flexibility, and data adaptation.


\section{Experimental results} 
\label{sec:results}
\begin{figure*}[htbp]
	\centering
	\includegraphics[scale=0.6]{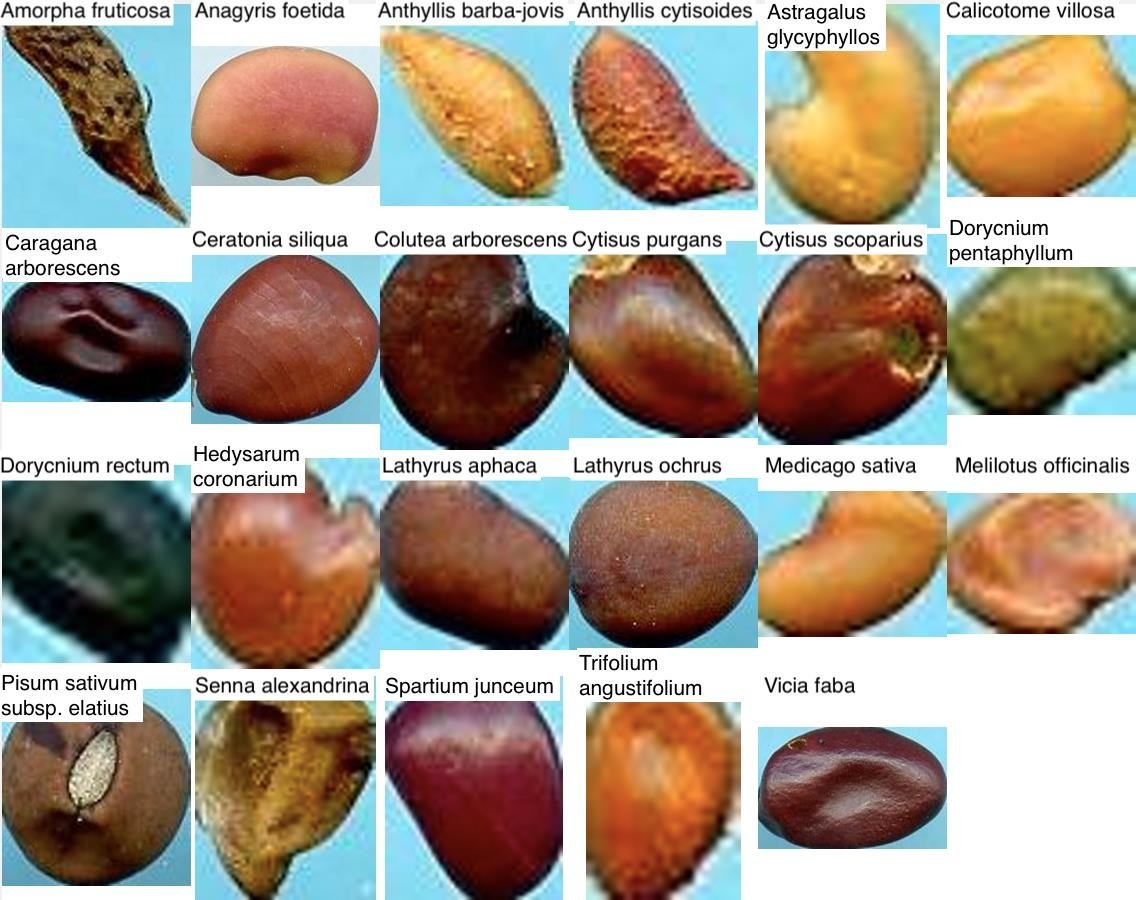}
	\caption{A sample of seed for each families present in the \emph{Fabaceae} dataset.}
	\label{Sardinia_samples}
\end{figure*}

We now describe the experimentations conducted to verify the correctness of the features extracted from \emph{SeedsAnalyser} using the classification plugin.
We selected the images containing the most numerous samples per families from the \emph{Fabaceae} database, for a total of 23 different ones: 
\emph{Amorpha}, \emph{Anagyris}, \emph{Anthyllis barba jovis}, \emph{Anthyllis cytisoides}, \emph{Astragalus glycyphyllos}, \emph{Calicotome}, \emph{Caragana}, \emph{Ceratonia}, \emph{Colutea}, \emph{Cytisus purgans}, \emph{Cytisus scoparius}, \emph{Dorycnium pentaphyllum}, \emph{Dorycnium rectum}, \emph{Hedysarum coronarium}, \emph{Lathyrus aphaca}, \emph{Lathyrus ochrus}, \emph{Medicago sativa}, \emph{Melilotus officinalis}, \emph{Pisum}, \emph{Senna alexandrina}, \emph{Spartium junceum}, \emph{Trifolium}, \emph{Vicia faba}, for a total of 1988 seeds.
A sample of each family is shown in Figure \ref{Sardinia_samples}, while Table \ref{Sardinia_tab} reports the number of samples for each family.

\begin{table}[htbp]
	\centering
	\small
	\hspace{0.05cm}
	\begin{tabular}{lc}
		\hline
		\textbf{families} & \textbf{Num. of samples} \\
		\hline
		\emph{Amorpha} & 51 \\
		\emph{Anagyris} & 29 \\
		\emph{Anthyllis barba jovis} & 51 \\
		\emph{Anthyllis cytisoides} & 29 \\
		\emph{Astragalus glycyphyllos} & 50 \\
		\emph{Calicotome} & 32 \\
		\emph{Caragana} & 36 \\
		\emph{Ceratonia} & 45 \\
		\emph{Colutea} & 42 \\
		\emph{Cytisus purgans} & 44 \\
		\emph{Cytisus scoparius} & 65 \\
		\emph{Dorycnium pentaphyllum} & 42 \\
		\emph{Dorycnium rectum} & 236 \\
		\emph{Hedysarum coronarium} & 208 \\
		\emph{Lathyrus aphaca} & 52 \\
		\emph{Lathyrus ochrus} & 46 \\
		\emph{Medicago sativa} & 116 \\
		\emph{Melilotus officinalis} & 176 \\
		\emph{Pisum} & 121 \\
		\emph{Senna alexandrina} & 194 \\
		\emph{Spartium junceum} & 109  \\
		\emph{Trifolium} & 183 \\
		\emph{Vicia faba} & 31 \\
		\hline
	\end{tabular}
	\caption{\emph{Fabaceae} dataset description.} 
	\label{Sardinia_tab}
\end{table}

\paragraph{Setup}
\label{par:setup}
As described in Sec. \ref{sec:seeds_analyser}, we have implemented and extracted three categories of handcrafted features from the seeds: morphological structure, texture information, and colour intensity values, for a total amount of 64 descriptors. Afterwards, we provided them as inputs to four different classification models, namely kNN, Naive Bayes, Random Forest, and Support Vector Machine, using our plugin \emph{SeedsClassifier}, based on Weka package tool \cite{Weka}. 

To ensure training set heterogeneity, we trained each classifier with 10-fold cross-validation, and for each case, we selected the model with the largest area under the ROC curve (AUC). 

\paragraph{Metrics}
\label{par:metrics}
The performance measures used to quantify each classification model's performance are specificity, sensitivity, and accuracy. 
The specificity (Spec) measures the proportion of negatives that are correctly identified (also called true negative rate). The sensitivity (Sen) measures the proportion of positives that are correctly identified (also called true positive rate).
The third measure is the accuracy (Acc), defined as the correctly labelled instances' ratio to the whole pool of instances. 
The measures are defined as follows:

\begin{equation}
	Accuracy = \frac{TP+TF}{TP+TF+FP+FN} ,
\end{equation} 

\begin{equation}
	Specificity = \frac{TN}{FP+TN} ,
\end{equation}

\begin{equation}
	Sensitivity = \frac{TP}{TP+FN} .
\end{equation}

TP, FP, TN, FN indicates True Positives, False Positives, True Negatives, False Negatives, respectively.

Moreover, as we face a multi-class imbalanced problem, we also applied three of the most common global metrics for multi-class imbalance learning to evaluate the classifier's performance~\cite{Alejo2013}. The used measures are the macro average geometric (MAvG), defined as the geometric average of the partial accuracy of each class, the mean F-measure (MFM) and the macro average arithmetic (MAvA) defined as the arithmetic average of the partial accuracies of each class. 

\begin{equation}
	MAvG = (\prod_{i=1}^J Acc_i)^{\frac{1}{J}} ,
\end{equation}

\begin{equation}
	MAvA = \frac{\sum_{i=1}^J Acc_i}{J} ,
\end{equation}

\begin{equation}
	MFM = \sum_{i=1}^J \dfrac{2 (\frac{Precision \cdot Sensitivity}{Precision+Precision})}{J} ,
\end{equation}

with Precision defined as follows: 
\begin{equation}
	Precision = \frac{TP}{TP+FP} .
\end{equation}

\paragraph{Results}
\label{par:results}
We performed several experiments for each classifier. In particular, we tested the descriptors categories both alone and in combination with the others to understand if there is the best descriptor category for this task. Finally, we use the \emph{SeedsClassifier} plugin to classify each category with the chosen classifiers.

The table \ref{tab:classicClassification} shows all the classification results on the analysed dataset.
In detail, the kNN classifier shows high results with colour features category alone, outperforming the remaining. Surprisingly, the combination of all categories does not reach the best metric results with this classifier.
Random Forest classifier substantially confirms the trend brought by the colour feature category. It outperforms every other combination, even against the remaining classifiers. However, the combination of all categories produced excellent results with the Random Forest model.
Naive Bayes and SVM classifiers produced satisfactory results using all categories, which result from the best solution for both classifiers. Contrary to kNN and Random Forest results, the colour category alone did not produce good results in these last two cases.
To sum up, the Random Forest classifier produced the best performance and is the only one to exceed 90\% both in metrics and in categories combination. It confirms its outstanding versatility in this twenty-three-class scenario. 
In summary, we can say that combining all the three feature categories produces excellent results in most cases and satisfactory on average.

\begin{table*}[tbhp]
    \centering
	\addtolength{\tabcolsep}{-0.1pt}
	\renewcommand{\arraystretch}{0.8}
	\caption{Performance results attained using HC descriptors and traditional classifiers.}
    \small
	\begin{tabular}{lllccccc}
		\toprule
		Classifier & Descriptor & Acc & Spec  & Sen & MAvG  & MFM & MAvA \\
		\midrule
		\multirow{7}{*}{\rotatebox[origin=c]{0}{kNN}}
		&Morphological & 20.95 & 23.18 & 17.10 & 9.94 & 18.59 & 23.18 \\
		&Texture & 16.41 & 20.52 & 16.97 & 10.91 & 16.36 & 20.52 \\
		&Colour & 80.54 & 76.59 & 74.70 & 74.72 & 75.15 & 76.59 \\
		&Morphological+Texture & 31.25 & 27.57 & 22.17 & 13.73 & 23.62 & 27.57 \\
		&Morphological+Colour & 45.13 & 41.63 & 33.47 & 27.08 & 35.34 & 41.63 \\
		&Texture+Colour & 68.61 & 62.33 & 58.44 & 57.74 & 59.73 & 62.33 \\
		&All & 71.68 & 69.54 & 63.02 & 65.60 & 65.17 & 69.54 \\ 
		\midrule
		\multirow{7}{*}{\rotatebox[origin=c]{0}{Naive Bayes}}
		&Morphological & 62.42 & 59.88 & 62.04 & 53.34 & 59.68 & 59.88 \\
		&Texture & 48.19 & 47.59 & 45.84 & 39.58 & 43.43 & 47.59 \\
		&Colour & 65.21 & 60.75 & 62.25 & 55.02 & 57.93 & 60.75 \\
		&Morphological+Texture & 76.81 & 73.00 & 75.10 & 68.51 & 72.89 & 73.00 \\
		&Morphological+Colour & 84.36 & 81.66 & 84.43 & 79.22 & 82.35 & 81.66 \\
		&Texture+Colour & 79.33 & 76.14 & 79.15 & 72.81 & 75.74 & 76.14 \\ 
		&All & 85.16 & 81.78 & 84.82 & 79.68 & 82.76 & 81.78 \\ 
		\midrule
		\multirow{7}{*}{\rotatebox[origin=c]{0}{RF}}
		&Morphological & 40.48 & 46.85 & 29.13 & 37.96 & 27.58 & 46.85 \\
		&Texture & 72.03 & 65.88 & 60.38 & 62.09 & 61.17 & 65.88 \\
		&Colour & 94.27 & 94.85 & 91.05 & 94.67 & 92.52 & 94.85 \\
		&Morphological+Texture & 89.64 & 90.67 & 81.96 & 90.09 & 83.79 & 90.67 \\
		&Morphological+Colour & 92.71 & 93.57 & 88.94 & 93.36 & 90.67 & 93.57 \\
		&Texture+Colour & 92.05 & 92.41 & 87.16 & 92.21 & 89.07 & 92.41 \\ 
		&All & 93.76 & 94.55 & 89.75 & 94.37 & 91.39 & 94.55 \\ 
		\midrule
		\multirow{7}{*}{\rotatebox[origin=c]{0}{SVM}}
		&Morphological & 79.83 & 80.26 & 71.33 & 79.46 & 73.23 & 80.26 \\
		&Texture & 29.09 & 46.28 & 21.13 & 34.11 & 16.91 & 46.28 \\
		&Colour & 78.74 & 75.28 & 67.83 & 72.42 & 67.88 & 75.28 \\
		&Morphological+Texture & 66.05 & 59.81 & 51.10 & 52.83 & 52.44 & 59.81 \\
		&Morphological+Colour & 83.60 & 83.10 & 76.70 & 81.65 & 78.88 & 83.10 \\
		&Texture+Colour & 84.51 & 84.99 & 77.73 & 84.15 & 78.81 & 84.99 \\
		&All & 85.66 & 83.85 & 78.88 & 82.56 & 80.58 & 83.85 \\ 
		\bottomrule
	\end{tabular}
	\label{tab:classicClassification}
\end{table*}

\begin{figure*}[tbhp]
	\centering
	\includegraphics[scale=0.60]{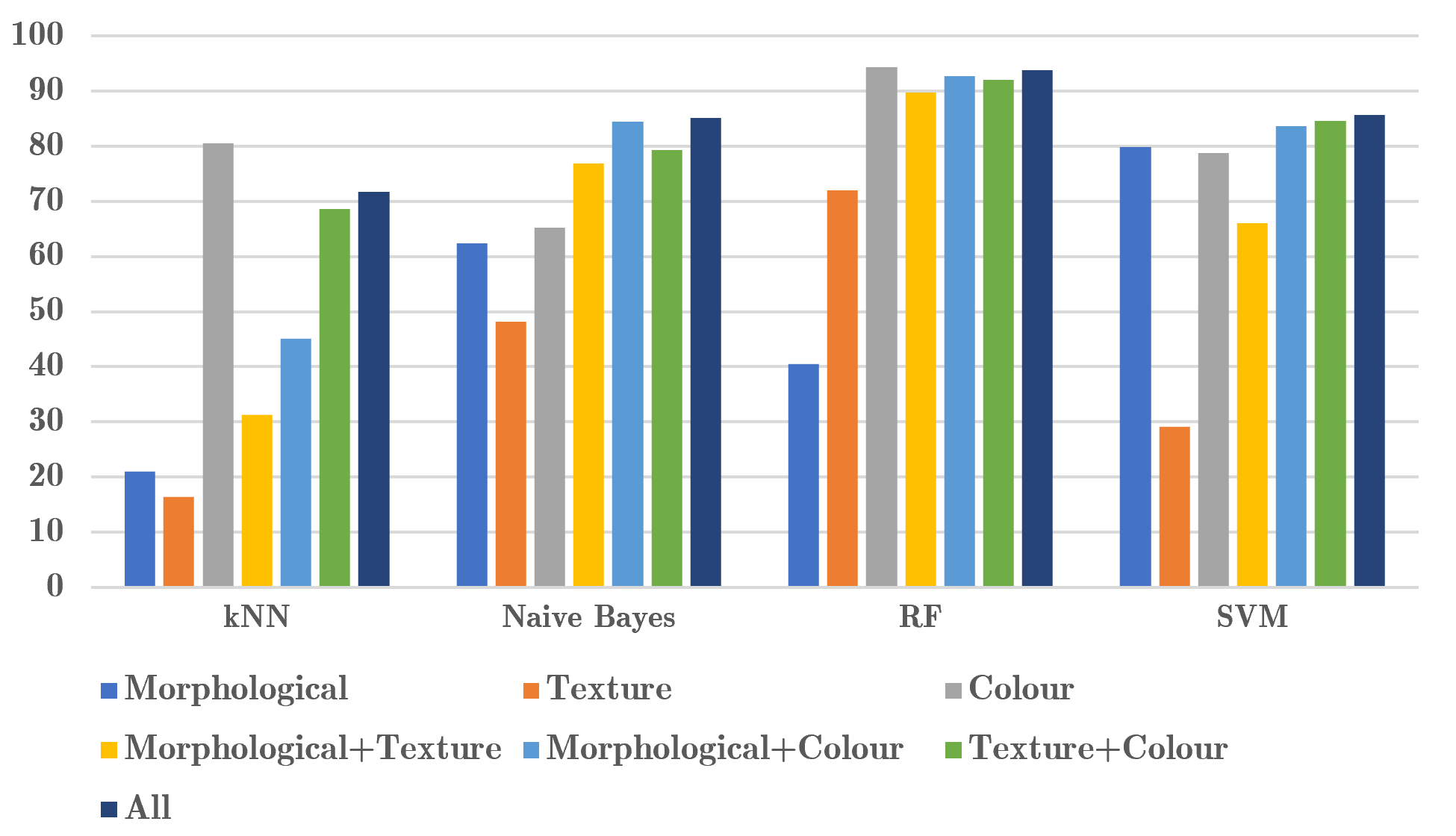}
	\caption{Graphical representation of the classifiers performance, as a function of each feature category.}
	\label{fig:g1}
\end{figure*}

\begin{figure*}[tbhp]
	\centering
	\includegraphics[scale=0.60]{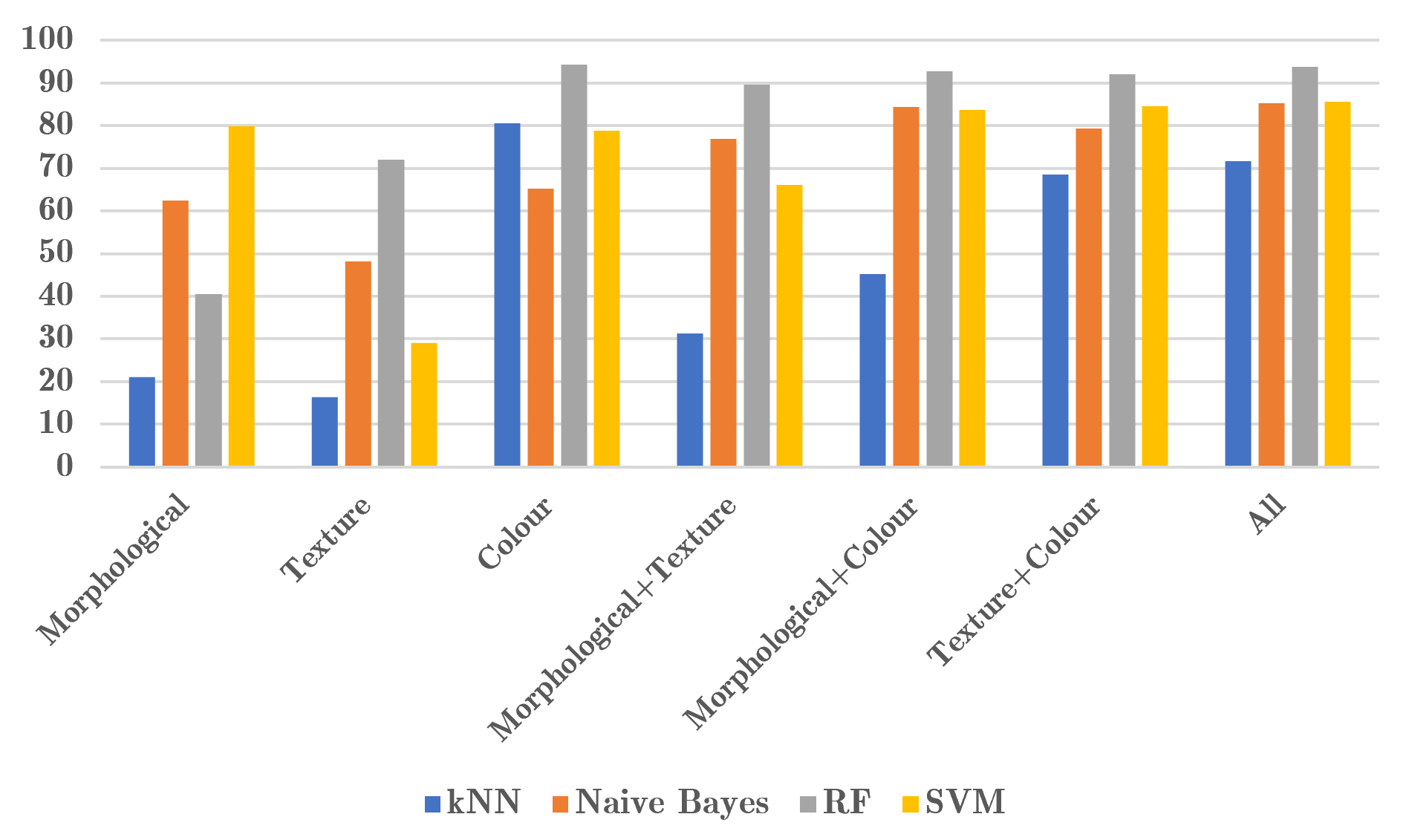}
	\caption{Graphical representation of the features performance, as a function of each classifier.}
	\label{fig:g2}
\end{figure*}

Fig. \ref{fig:g1} shows that the Random Forest classifier obtained the best performances for all the descriptor categories considered, except for the morphological one, in which the SVM classifier outperforms the others. 
As regards the descriptors categories, Fig. \ref{fig:g1} shows the relevance of the colour features in this context. Indeed, they, both alone and combined with morphological or textural, produced an increasing accuracy for all the classifiers tested. In some cases, they obtained values close or higher than the combination of all the features categories, as shown in Fig. \ref{fig:g2}.

Finally, we performed several comparisons with deep learning-based approaches. In table \ref{Table_classification_CNN} we report the performance results obtained using the best classical descriptors with the relative classification model and some different CNNs. As expected, CNNs achieved better performances than traditional methods. Nevertheless, they presented a much longer training time than that required by any approach based on traditional descriptors and machine learning models. With just a slight loss of accuracy, our tool establishes itself as the best solution for practical and immediate use of seed image analysis. 

\begin{table*}[tbhp]
	\centering
	\caption{Results using the best classic descriptors and relative models and some CNNs. The results are sorted by the training time, indicated by the Time column.}
	\label{Table_classification_CNN}
	\begin{tabular}{lccccccr}
		\toprule
		Method    &  Acc  & Spec  &  Sen  & MAvG  &  MFM  & MAvA  & Time\\ \midrule
		kNN & 80.54 & 76.59 & 74.70 & 74.72 & 75.15 & 76.59 & 8 sec \\
		SVM  & 85.66 & 83.85 & 78.88 & 82.56 & 80.58 & 83.85 & 18 sec \\
		Random Forest & 94.27 & 94.85 & 91.05 & 94.67 & 92.52 & 94.85 & 29 sec \\
		Naive Bayes  & 85.16 & 81.78 & 84.82 & 79.68 & 82.76 & 81.78 & 64 sec \\
		SeedNet & 97.47 & 99.88 & 96.81 & 96.60 & 96.98 & 96.81 & 12 min \\
		ShuffleNet & 96.46 & 95.90 & 94.37 & 95.57 & 94.55 & 95.90 & 18 min \\
		GoogLeNet & 95.45 & 95.06 & 93.47 & 94.67 & 93.16 & 95.06 & 21 min \\
		SqueezeNet & 95.96 & 95.75 & 94.71 & 95.13 & 94.84 & 95.75 & 23 min \\
		MobileNetV2 & 93.94 & 93.16 & 91.51 & 92.67 & 91.85 & 93.16 & 33 min \\
		ResNet50 & 96.46 & 94.44 & 94.98 & 96.15 & 95.20 & 96.44 & 34 min \\
		AlexNet & 93.43 & 91.08 & 91.36 & 90.15 & 90.51 & 91.08 & 74 min\\
		VGG16 & 95.96 & 94.82 & 94.86 & 94.22 & 94.25 & 94.82 & 224 min \\
		InceptionV3 & 96.46 & 95.99 & 94.81 & 95.74 & 95.02 & 95.99 & 290 min \\
	 \bottomrule
	\end{tabular}
\end{table*}


\section{Conclusions}
\label{sec:conclusions}
We presented a software that performs an image analysis by feature extraction and classification from images containing seeds through a brand new unique, and easy-to-use framework. In detail, we propose two \emph{ImageJ} plugins, one able to extract morphological, textural and colour characteristics from images of seeds, and another one to classify the seeds into categories by using the extracted features. 
Moreover, we analysed and reported the performances of several categories of descriptors for seed images with four different classifiers, using an image database containing 3,386 samples of 120 plant families belonging to the \emph{Fabaceae} family. 
In general, some aspects can strongly influence both the feature extraction and the classification phases. Foremost, the quality of the original images to process can produce some artefacts in the segmentation phase. Secondly, the preprocessing step, such as the background cleaning, the spacing of the seeds during the acquisition, and the size of the seeds present in the images, need to be verified to consider only valid regions. Finally, the dataset represented a class imbalance problem.

The experiments carried out showed some interesting trends. The colour feature category alone produced the best results in every metric, either using kNN and Random Forest classifiers. However, apart from Naive Bayes and SVM results in which they were the best, the combination of all the three categories produced excellent results on average. Finally, the Random Forest was the only one to outrun 90\% both in metrics and in categories combination, showing its excellent versatility.
The comparison with deep learning-based approaches has shown how, although slightly decreasing the accuracy, all the classic methods based on machine learning and hand-crafted descriptors are much faster with acceptable performance. Therefore, the proposed tool lends itself to be used for quick and practical use of seed image analysis, especially because of its speediness both in training and testing.

As a future direction, thanks to the promising results obtained by the proposed tool, we aim to improve them further by investigating the possibility of using a feature selection step and optimising the parameters that characterise the chosen classifiers in the training phase. We also plan to consider neural network features extraction and compare them with the traditional ones. 
Finally, we would like to extend our approach to distinguish among seeds' genus and variety.


\newpage
\newpage
\section*{Abbreviations}
The following abbreviations are used: 

	\begin{tabular}{l | l}
        DPI     &   Dots Per Inch                       \\
    	CNN     &   Convolutional Neural Network        \\
    	HC      &   Handcrafted                         \\
    	Acc     &   Accuracy                            \\
    	Spe     &   Specificity                         \\
    	Sen     &   Sensitivity                         \\
    	MAvG    &   Macro Average Geometric             \\
    	MFM     &   Mean F-Measure                      \\
    	MAvA    &   Macro Average Arithmetic            \\
    	SVM     &   Support Vector Machine              \\  
    	KNN     &   K Nearest Neighbour                 \\
    	RF      &   Random Forest                       \\
    	GLCM    &   Gray-Level Co-Occurrence Matrix     \\    
    \end{tabular}

\section*{Conflict of interest}
The authors declare that they have no conflict of interest.

\bibliographystyle{spbasic}      
\bibliography{bib}

\end{document}